\documentclass[10pt,twocolumn,letterpaper]{article}
\usepackage[accsupp]{axessibility}  
\usepackage{iccv}
\usepackage{times}
\usepackage{epsfig}
\usepackage{graphicx}
\usepackage{amsmath}
\usepackage{amssymb}
\usepackage{setspace}
\usepackage{amsmath}
\usepackage{algorithmic}
\usepackage{multirow}
\usepackage{threeparttable}
\usepackage{bm}
\usepackage{xspace}
\usepackage{array}
\usepackage{diagbox}

\usepackage[pagebackref=true,breaklinks=true,letterpaper=true,colorlinks,bookmarks=false]{hyperref}

\iccvfinalcopy 

\ificcvfinal\fi

\begin{document}

\title{Pre-training-free Image Manipulation Localization through Non-Mutually Exclusive Contrastive Learning}

\author{Jizhe Zhou$^{1,4}$,
Xiaochen Ma$^{1,4}$,
Xia Du$^{2,4}$,
Ahmed Y.Alhammadi$^{3,4}$,
Wentao Feng$^{1,4}$\thanks{Wentao Feng is the corresponding author, contact through mail address: Wtfeng2021@scu.edu.cn}\\
$^{1}$\normalsize{College of Computer Science, Sichuan University}\\
$^{2}$\normalsize{School of Computer and Information Engineering, Xiamen University of Technology}\\
$^{3}$\normalsize{Strategy Affairs Office, Mohamed Bin Zayed University for Humanities}\\
$^{4}$\normalsize{Engineering Research Center of Machine Learning and Industry Intelligence, Ministry of Education of China}
}
\maketitle
\ificcvfinal\thispagestyle{empty}\fi

\begin{abstract}
Deep Image Manipulation Localization (IML) models suffer from training data insufficiency and thus heavily rely on pre-training. We argue that contrastive learning is more suitable to tackle the data insufficiency problem for IML. Crafting mutually exclusive positives and negatives is the prerequisite for contrastive learning. However, when adopting contrastive learning in IML, we encounter three categories of image patches: tampered, authentic, and contour patches. Tampered and authentic patches are naturally mutually exclusive, but contour patches containing both tampered and authentic pixels are non-mutually exclusive to them. Simply abnegating these contour patches results in a drastic performance loss since contour patches are decisive to the learning outcomes. Hence, we propose the Non-mutually exclusive Contrastive Learning (NCL) framework to rescue conventional contrastive learning from the above dilemma. In NCL, to cope with the non-mutually exclusivity, we first establish a pivot structure with dual branches to constantly switch the role of contour patches between positives and negatives while training. Then, we devise a pivot-consistent loss to avoid spatial corruption caused by the role-switching process. In this manner, NCL both inherits the self-supervised merits to address the data insufficiency and retains a high manipulation localization accuracy. Extensive experiments verify that our NCL achieves state-of-the-art performance on all five benchmarks without any pre-training and is more robust on unseen real-life samples. \href{https://github.com/Knightzjz/NCL-IML}{https://github.com/Knightzjz/NCL-IML}.
\end{abstract}

\section{Introduction}
\label{sec:intro}
\begin{figure}[ht]
\centering
\includegraphics[width=1.0\columnwidth]{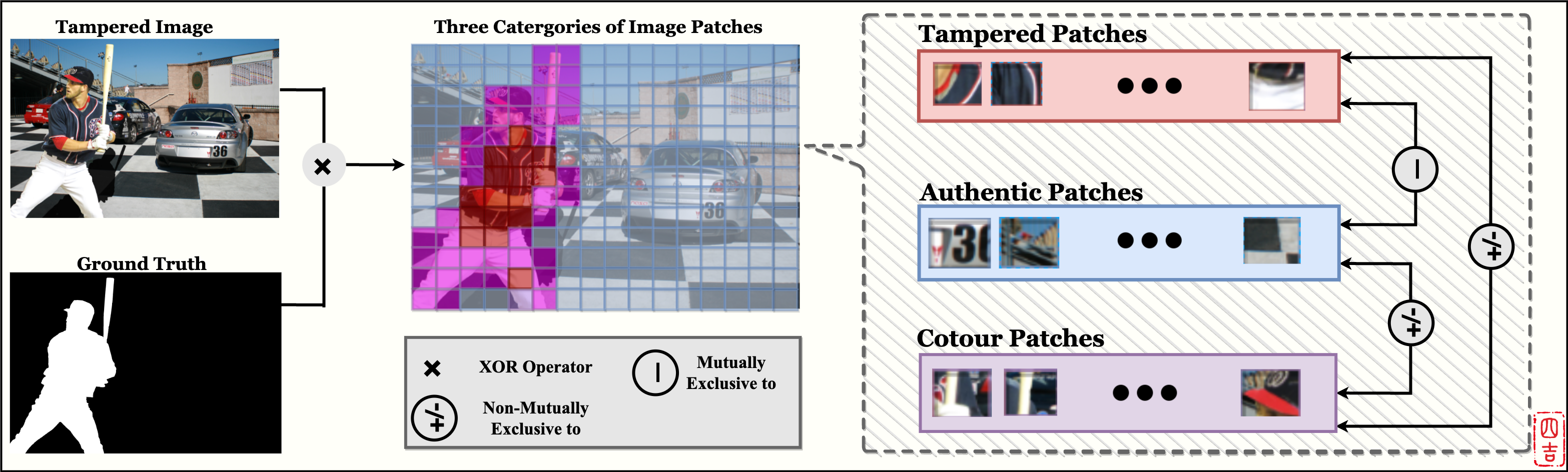}
\caption{Three categories of patches in a manipulated image and the non-mutually exclusive relations among them. There are three categories of patches in a manipulated image: tampered, authentic, and contour patches. In the middle picture, we depict them accordingly with red, blue, and purple squares. Only tampered and authentic patches are mutually exclusive. When the decisive contour patches are involved, non-mutually exclusivity occurs in contrastive learning. Best viewed in color.}
\label{fig:overall}
\vspace{-1em}
\end{figure}
Thrilling advances in media techniques grant us easier and easier access to manipulate images. Image Manipulation Localization (IML) is then indispensable for defensive information forensics and is heavily invested by the information security industry. Today, data insufficiency is the most prominent issue in building deep IML models. As dense annotations and expertise for tamper identification are exorbitant, public datasets for IML are all tiny-sized (with a few hundred to a few thousand images) and severely insufficient for training deep CNNs. Consequently, major deep IML methods carry out pre-training on additional large-scale datasets. 
\par
In general, pre-training of IML models relies on synthesized datasets. On the one hand, synthesized datasets vanish the high labeling costs and pre-training on synthesized datasets refrains from overfitting. On the other hand, employing synthesized datasets to conduct pre-training impedes fair comparisons among models and even jeopardizes the model generalizability. Pre-training is crucial to the model performance, and for fair comparisons, models of the same task commonly practice their pre-training on the same dataset. However, synthesized pre-training datasets for IML models are strikingly different in annotation quantity and quality. For instance, ManTra-Net~\cite{Wu2019ManTraNet} grounds on a self-collected, pixel-wise labeled dataset of 102,028 images and 385 manipulation types for pre-training; RGB-N~\cite{zhou2018RGB-N} employs a randomly synthesized dataset of more than 42,000 images; BusterNet~\cite{BusterNet} entails a synthesized dataset of 100,000 copy-moved images for pre-training; MVSS~\cite{dong2022mvss} adopts a synthesized dataset of 8,4000 images. Faithful evaluations for models pre-trained on different synthesized datasets become impossible. Moreover, unlike real tampered images, these naively synthesized images severely lack elaborate post-processing to cover their manipulation traces or artifacts~\cite{chen2021image, wang2022objectformer,dong2022mvss}. In other words, the sampling process of synthesized datasets is biased from the sampling process of manual build datasets~\cite{zhou2020personal, zhou2020privacy}. A model learned on such a dataset with sampling bias is short in generalizability, and measuring this mode on tiny-sized, non-homologous benchmarks cannot fully disclose its poor performance under real cases.   
\par 
To address the insufficient data problem without introducing such a tricky pre-training strategy, we advocate adopting contrastive learning in IML. On the one hand, self-supervised contrastive learning can yield massive contrastive pairs from real tampered images. These contrastive pairs boost the training sample number by at least one or two orders of magnitude without causing sampling bias or unfaithful evaluations. On the other hand, manipulation leaves artifacts in images, and artifacts cause feature discrepancies between tampered and authentic regions. This is the essential clue for identifying tampered areas by human experts. The contrastive learning objective explicitly follows this clue and reveals the vital feature discrepancies by encouraging the compactness between positive pairs and the margin between negative pairs.
\par 
Although recent researches suggest pixel-level contrastive learning for pixel predictions~\cite{xie2021propagate}, patch-wise contrastive learning is still more suitable for IML. Because manipulations rarely happen pixel-by-pixel, the patch-level features are proven to be outstanding in characterizing manipulation traces or artifacts~\cite{ma2023iml}. Thus, in our method, positives and negatives are naturally the tampered and authentic image patches of pure tampered or authentic pixels. Image patches are in a fixed size, but the manipulated regions are arbitrarily shaped and sized. As shown in the middle picture of Figure~\ref{fig:overall}, when sampling along the contour of manipulated regions, tampered and authentic pixels are inevitably mingled within one image patch. Then, we have the third patch, contour patches. Apparently, contour patches are neither mutually exclusive to tampered patches nor authentic patches. Conventional contrastive learning designed to handle the mutually exclusive relation between binary sets will then malfunction under such a trilateral, non-mutually exclusive circumstance. However, simply discarding the contour patches and merely employing the tampered and authentic patches to conduct contrastive learning is not feasible. Previous studies~\cite{NOI2009, splice2017MFCN, BLK2009, ma2023iml} show that artifacts assemble along the borders of tampered areas. Therefore, discarding contour patches means throwing away samples with the richest artifacts' information. Besides, contour patches are the hard positives or negatives in contrastive learning since they contain both tampered and authentic pixels at the same time. Hard samples are decisive to the contrastive learning outcomes. Discarding contour patches also eliminates most of the hard samples in contrastive learning. In short, we are facing such a dilemma: the existing contrastive learning paradigm is incompatible with the non-mutually exclusive contour patches, but learning without contour patches results in a significant performance gap, and learning without the contrastive paradigm leads to model generalization and evaluation issues. Therefore, a brand-new learning framework that follows the contrastive learning paradigm and copes with non-mutual exclusivity is the key to saving IML models from this dilemma.

\par
Hence, we propose the~\textit{Non-mutually exclusive Contrastive Learning} (NCL) framework. Every contour patch is partial-tampered and partial-authentic. Therefore, we can regard a contour patch as a hard positive in contrastive learning if we only count its tampered part. Likewise, this counter patch can be simultaneously regarded as a hard negative if only its authentic parts are counted. That is, a contour patch can be transferred into a hard positive or a hard negative referring to its partial information. Following this role-switching characteristic, we constructed a pivot structure with dual branches on the shallow layers of the backbone to squeeze the positive and negative parts accordingly from the contour patches. The name of the pivot indicates that it switches contour patches between the role of hard positives and hard negatives to constitute contrastive pairs. Thus, the trilateral, non-mutually exclusive contrast among tampered patches (positives), authentic patches (negatives), and the contour patches is then disentangled into three binary, mutually exclusive, contrastive pairs of~\textit{\{positive, negative\}},~\textit{\{positive, hard negative\}},~\textit{\{negative, hard positive\}}. The NCL loss is the sum of the three pair-wise contrastive losses. In addition, the pivot structure corrupts the spatial correlation among contour patches. Therefore, on the decoder side, we devise the \textit{pivot-consistent loss} with auxiliary classifiers to ensure the pixel-wise spatial relations are captured and preserved by the deeper layers of the encoder. 
\par 
We train our NCL-based method from scratch without additional datasets or pre-training stages. With only 5-10\% of the total training data compared with pre-training-based methods, our model outperforms current pre-training-based approaches on all five public IML benchmarks. Despite this, deep CNNs are prone to overfitting on such small public benchmarks. Therefore, we further use non-homogeneous training and testing datasets to examine model generalization ability. The results verify that NCL endows our IML model with better localization accuracy and robustness. Last but not least, similar to contrastive learning, NCL also holds the plug-in merit. Regardless of backbone architecture, NCL functions well.

\par
In summary, our main contributions are quad-folded:
\begin{itemize}
    \item \textbf{Free of Additional Data}. To the best of our knowledge, we are the first work bringing contrastive learning in IML to address the insufficiency of training data and drawbacks caused by pre-training.
    \item \textbf{Non-Mutually Exclusive Contrast}. As far as we know, we are also the first to handle non-mutual exclusive, trilateral relations through contrastive learning. Our Non-mutually exclusive Contrastive Learning (NCL) framework can serve other tasks like semantic segmentation or fine-grained object detection.
    \item \textbf{Top Benchmark Performance}.  Our method uses less and inferior training data but achieves state-of-the-art performances as well as the top model generalization ability on all five public benchmarks. 
 \item \textbf{Plug-in Merit}. Our method functions under both CNN and Transformer-styled backbones. Backbone selection will not break the integrity of NCL.   
   \end{itemize}

\section{Related Work}
\label{sec:related}
\textbf{Image Manipulation Localization}. 
Prior IML methods seek for pre-training strategy, hand-crafted features, and the self-adversarial paradigm to solve the data insufficiency issue. As discussed in the previous section, methods~\cite{wang2022objectformer,dong2022mvss, Deng2020CVPR, zhou2020personal} involving hand-crafted features or pre-training mechanisms are not the proper solution for the insufficient data issue. We here view the other Generative-Adversarial Network (GAN) based methods in IML. GAN-based solutions~\cite{splice2019MAG, DOA-GAN, zhuo2021self} also reach state-of-the-art performances without additional datasets. However, primary GAN-based methods are sensitive to the manipulation types.~\cite{DOA-GAN} only works on copy-moved images;~\cite{splice2019MAG} is practical merely for splicing manipulation. Our most related study is the self-adversarial GAN~\cite{zhuo2021self}. They also noticed the drawbacks of pre-training and built a self-adversarial training strategy in a dual-attention GAN to localize forged regions precisely. However, GAN-based methods do not explicitly follow the clue of image manipulation, which is the discrepancies between tampered and authentic regions, thereby undermining the model interpretability. Moreover, the generated training samples are still different from the real ones, thereby undermining model performances on real-life images. Our proposed NCL reveals the essential tamper-caused feature differences as well as boosts the number of real training samples.
\par
\textbf{Contrastive Learning}. 
Contrastive learning~\cite{chen2020improved} is emerging and fast developing in self-supervised and unsupervised visual representation areas. Conventional contrastive learning is commonly applied in tasks whose problem space is bisected. Binary and mutually exclusive relations are the fundamental assumption to apply existing contrastive learning. This is why existing contrastive IML models~\cite{Self-Consistency2018eccv, mayer2019forensic, spliceDT2017} only conduct comparisons on images rather than image patches. As far as we know, current studies can only handle binary (similar or dissimilar) contrasts~\cite{he2020momentum, robinson2020contrastive}. Our NCL extends the contrastive learning paradigm into non-mutually exclusive relations among trilateral sets, thereby retains the information-rich contour patches and gains surpassing performances in the IML task. 

\section{Method}
\label{sec:method}

\begin{figure*}[!t]
\centering
\includegraphics[width=1.0\textwidth]{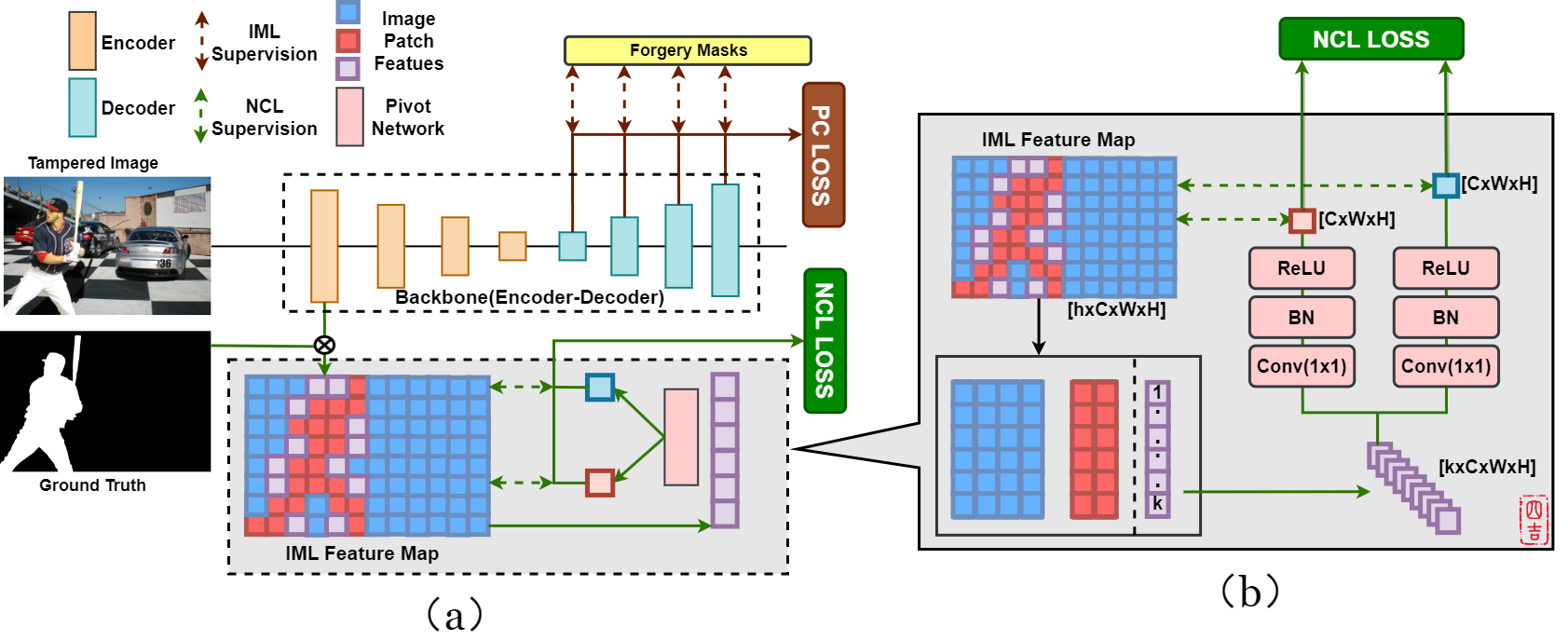}
  \caption{(a): General network structure of our NCL framework. (b): Detailed Pivot network structure. Green-colored arrows signify the flow of conducting non-mutually exclusive contrasts through Pivot network and then generate Non-mutually Exclusive Contrastive Learning (NCL) loss. Ocher-colored arrows indicate the flow to generate Pivot-Consistent (PC) loss. Feature map output by the first encoder block is point-wise classified into tamper (red), authentic (blue), and contour (purple) features according to ground truth. Forgery masks in yellow rectangular are the ground truth in different sizes. Feature sizes are enclosed by brackets.}
\label{fig:onecol}
\end{figure*}

\subsection{Basic Encoder-Decoder Structure}
\par

We adopt DeepLabV3+~\cite{DeepLabV3+} as the basic encoder-decoder structure of our IML model since it has been adopted by many other IML models as the baseline~\cite{gao2021tbnet, dong2022mvss}. Do notice, the base mode selection or the backbone selection will affect the efficacy of our NCL. Thus, the encoder backbone in Figure~\ref{fig:onecol} is ResNet101~\cite{ResNet} blocks with atrous convolution in the last few blocks. The Atrous Spatial Pyramid Pooling (ASPP) block is likewise applied. Afterward, the encoded feature of size (64$\times$ 64) is passed to the decoder. The decoder adopts two upsampling modules. The encoder output is twice upsampled by a factor of 4. In short, our basic encoder-decoder applies the same network structure and training settings as the DeepLabV3+ model.

\subsection{Non-Mutual Exclusive Contrastive Learning}
\textbf{Problem Formulation}. For conventional contrastive learning, define the problem domain as the universal set $\mathbb{U}$. Like the conventional contrastive learning section in Figure~\ref{fig:overall} shows, we have the set of positives $\mathbb{P}$, and set of negatives $\mathbb{N}$, where:
\begin{equation}
\begin{aligned}
	&\mathbb{P} \cup \mathbb{N} = \mathbb{U} \\
	&\mathbb{P} \cap \mathbb{N} = \emptyset
	\end{aligned}
	\label{eq1}
	\end{equation}
$\emptyset$ indicates the mutual exclusivity between positives and negatives. Mark $p$ as one tampered image patch, which is one element of $\mathbb{P}$. For $\forall p \in \mathbb{P}$, we further denote $p_j \in \mathbb{P}, p_j \neq p$; and $n_i \in \mathbb{N}$. Then, the conventional contrastive learning objective is:
\begin{equation}
	 \arg\max_{f}\{\sum_{i,j}\phi(f(p), f(n_i))-\phi(f(p), f(p_j))\}
\label{eq2}
\end{equation}
$f(\cdot)$ is the learned feature representation for an image patch. $f(p_j)$ and $f(n_i)$ are the red and blue cubes in the IML feature map in Figure~\ref{fig:onecol}. $\phi(\cdot,\cdot)$ represents the measured distance, namely the similarity, between two feature vectors. Notations are unified throughout this paper, where sets of image patches are denoted by upper-case letters, image patches are represented by lower-case letters, and $f(\cdot)$ function is the learned feature representation of an image patch.
\par 
However, for the NCL illustrated Figure~\ref{fig:overall}, we have:
\begin{equation}
\begin{aligned}
    &\mathbb{N} \cup \mathbb{P} \cup\mathbb{C} =\mathbb{U}\\
    &\mathbb{P} \cap \mathbb{{N}} =\emptyset;
    \mathbb{C} \cap \mathbb{{N}}= \mathbb{C^-};
    \mathbb{C} \cap \mathbb{{P}}= \mathbb{C^+}
    \end{aligned}
    \label{eq3}
\end{equation}
$\mathbb{C}$ is the set of all contour patches. $\mathbb{C^+}$ and $\mathbb{C^-}$ are denoted as $\mathbb{C}$'s intersections with the positives set and negatives set. Meaning the positive and negative pixels mingled in the contour patches. For contrastive learning, positive pairs can be easily formed by finding another element in the same set. According to~(\ref{eq1}) and ~(\ref{eq2}), the empty intersection implies how to form the important negative pairs. Therefore, we first revise~(\ref{eq3}) into the exact same format with~(\ref{eq1}). With little tricks, we can have:  
\begin{equation}
\begin{aligned}
    &\mathbb{N} \cup \mathbb{P} \cup\mathbb{C} =\mathbb{U}\\
    &\mathbb{P} \cap \mathbb{{N}} =
    \mathbb{C^+} \cap \mathbb{{N}}=
    \mathbb{C^-} \cap \mathbb{{P}}=\emptyset
    \end{aligned}
    \label{eq4}
\end{equation}
Then, according to~(\ref{eq1}), we now can transfer the non-mutually exclusive contrast written in~(\ref{eq3}) into three binary contrasts between ($\mathbb{P} \cap \mathbb{{N}}$), ($ \mathbb{C^+} \cap \mathbb{{N}}$), and ($\mathbb{C^-} \cap \mathbb{{P}}$). To carry out the three pair-wise comparisons, we need to first find out $\mathbb{C^+}$ and $\mathbb{C^-}$ defined in~(\ref{eq3}). Also $\mathbb{C^+}$ and $\mathbb{C^-}$ are patch fragments or pixels. The basic encoder network cannot yield features for patch fragments. So, we design the pivot network to directly use contour patches as the input and generate feature representations for $\mathbb{C^+}$ and $\mathbb{C^-}$. That is, the pivot network switches the role of contour patches by learning two mapping function between $(\mathbb{C}, \mathbb{C^+})$ and $(\mathbb{C}, \mathbb{C^-})$. Naturally, the pivot network should own two similar branches with the same input.

\par 
\textbf{Pivot Network}.
Before building the detailed layouts for the pivot network, we need to further consider the input of the pivot network. Training the pivot network also requires adequate contour patches. But, if we select a small patch size to generate more contour patches. The small patch size leads to a small number of pixels in one image patch. Then, some elements in $\mathbb{C^+}$ or $\mathbb{C^-}$ may contain a trivial number of pixels and are improper for training the pivot network. Hence, in a single image, we concatenate all contour patch features into one entire embedding $\mathfrak{p}$, and send $\mathfrak{p}$ as the input of the pivot network to ensure the learning outcomes are significant enough for comparison. In Figure~\ref{fig:onecol}, this concatenation assembles purple cubes into a strip of size $(k \times C \times W \times H)$. $k=card({\mathbb{C}})$. $C, W, H$ are the channel, height, and width of one contour feature. $card()$ denotes the cardinality or the number of elements in a set $\mathbb{C}$. On the one hand, with $k=card(\mathbb{C})$, we concatenate the contour patch features into a single vector $(k \times C \times W \times H)$. This vector aggregates contour patch features within an entire image to address the model inefficiency when a few contour patches exist. On the other hand, the Pivot network flattens this $(k \times C \times W \times H)$ vector into a fix-sized $(1 \times C \times W \times H)$ vector. This further helps to deal with the varying size of $k$ in feature processing.
\par
The detailed structure of the Pivot network is depicted in Figure~\ref{fig:onecol} (b) through pink rectangles and green arrows.
\par 
Then, we design two symmetrical branches for our pivot network. These branches share the same input and have the same structure. $\mathfrak{p}$ is first process by the $(1 \times 1)$ convolution. This $(1 \times 1)$ convolution kernel flattens $\mathfrak{p}$ into the shape of $(1 \times C \times W \times H)$. Moreover, this $(1 \times 1)$ kernel projects $\mathfrak{p}$ into a latent Hilbert space $\mathcal{H}: \mathbb{R}^{C \times W \times H}$, where $f(p_j)$ and $f(n_i)$ settle and the similarity between features can be uniformly measured by $\phi(\cdot, \cdot)$. BN and ReLU are the batch normalization and ReLU activation layers. 
\par 
The pivot network constructs the reflection $f(\cdot)$ between the input set $\mathbb C, (c\in \mathbb C)$ and output sets $\mathbb C^+, (c^+ \in \mathbb C^+)$ and $\mathbb C^-, (c^- \in \mathbb C^-)$. So, $f(\cdot)$ are expected to satisfy:\\
\indent (1).$\mathbb C^+$ and $\mathbb C^-$ benefit the IML accuracy;\\
\indent (2).$\mathbb C^+$ and $\mathbb C^-$ are smooth manifolds to ensure the back-propagation of NCL loss. Since $\mathbb C$ is a smooth manifold (limited Euclidean space), $f(\cdot)$ should be a bijection;\\
\indent (3).No information loss after the reflection. Meaning we can assemble $c^+$ and $c^-$ back to $c$ through some binary operation $(\cdot)$; $c^+ \cdot c^- = c$, $c^+ \cdot c = c$, $c^- \cdot c = c.$\\
\indent Thus, we can have a Group $(G,{\cdot})$, where $G=\mathbb C^+\cup \mathbb C^-$. $G$ is a Lie group because:\\
$\square$ \hspace{0.1cm}The group inverse $G \xrightarrow[]{} G$ is smooth according to (2).\\
$\square$ \hspace{0.1cm}The group product $G\times G \xrightarrow[]{} G$ is smooth due to (3).
\par 
Therefore, the output of the pivot network ($c^+$ and $c^-$) are Lie group elements. We then take the pivot network as a smooth mapping function and borrow the $\mathfrak{se}$ notation from the Lie group. We write the output of the two branches as $\mathfrak{se}^+(\mathfrak{p})$ and $\mathfrak{se}^-(\mathfrak{p})$. $\mathfrak{se}^+(\cdot)$ and $\mathfrak{se}^-(\cdot)$ just signify the feature transformation function learned by Pivot network; we cannot assure they are differential manifolds. $\mathfrak{se}^+(\mathfrak{p})$ and $\mathfrak{se}^-(\mathfrak{p})$ are the light red and light blue cubes yielded in Figure~\ref{fig:onecol} (b). The sets of $\mathfrak{se}^+(\mathfrak{p})$ and $\mathfrak{se}^-(\mathfrak{p})$ are the desired $\mathbb{PI^+}$ and $\mathbb{PI^-}$. An intuitive explanation of  $\mathfrak{se}^+(\mathfrak{p})$ and $\mathfrak{se}^-(\mathfrak{p})$ is they are special positive and negative features squeezed from the generated feature $\mathfrak{p}$ by Pivot network; while common positive and negative features are generated by the backbone network according to physically-existing image patches. From this point of view, the Pivot network swings the role of pivot between positives and negatives like a pendulum.
\par 
Based on $f(\cdot)$ and $\phi(\cdot, \cdot)$ in $\mathcal{H}$, $\mathfrak{se}^+(\cdot)$ and $\mathfrak{se}^-(\cdot)$, we formulate the NCL learning objective as:
\begin{equation}
\begin{aligned}
&\arg\max_{f, \mathfrak{se}^+, \mathfrak{se}^-}\{\sum_{i,j}\phi(f(p), f(n_i))-\phi(f(p), f(p_j))\}+	 \\
     &\{\sum_{i,j}\phi(\mathfrak{se}^+(\mathfrak{p}), \mathfrak{se}^-(\mathfrak{p}))-\phi(\mathfrak{se}^+(\mathfrak{p}), {f(p_j)})\}+ \\
     &\{\sum_{i,j}\phi(\mathfrak{se}^+(\mathfrak{p}), \mathfrak{se}^-(\mathfrak{p}))-\phi(\mathfrak{se}^-(\mathfrak{p}), f({n_i}))\}	
\end{aligned}	
\label{eq5}
\end{equation}

\par 
\textbf{Non-Mutually Exclusive Contrast Loss}.
We indeed can construct NCL loss function according to~(\ref{eq5}). But, as the pivot network yields one $\mathfrak{se}^+(\mathfrak{p})$ and one $\mathfrak{se}^-(\mathfrak{p})$ for each manipulated image, $\phi(\mathfrak{se}^+(\mathfrak{p}), \mathfrak{se}^-(\mathfrak{p}))$ is independent from summing parameter $i,j$ and becomes a constant amid the loss accumulation process. Such a constant undermines the diversity of contrastive pairs. Hence, we make minor substitutions in the construction of positive pairs and further refine~(\ref{eq5}) as:
 \begin{equation}
\begin{aligned}
&\arg\max_{f, \mathfrak{se}^+, \mathfrak{se}^-}\{\sum_{i,j}\phi(f(p), f(n_i))-\phi(f(p), f(p_j))\}+	 \\
     &\{\sum_{i,j}\phi(\mathfrak{se}^+({\mathfrak{p}}), f({n_i}))-\phi({\mathfrak{se}^+(\mathfrak{p})}, {f(p_j)})\}+ \\
     &\{\sum_{i,j}\phi(\mathfrak{se}^-(\mathfrak{p}), f({p_j}))-\phi(\mathfrak{se}^-(\mathfrak{p}), f({n_i}))\}	
\end{aligned}	
\label{eq6}
\end{equation}
Through our pivot network, in~(\ref{eq6}), NCL reforms the non-mutually exclusive relation among trilateral image patches into three mutual-exclusive, pair-wise, binary comparisons connected by $``+"$. This is drawn by the NCL supervision in Figure~\ref{fig:onecol}. For simplification, we assign $p$ a subscript by letting $p=p_m$; mark $e_{x}^{y}=\exp(f(x), f(y)) / \tau$, $e^-_{x}=\exp(\mathfrak{se}^-(\mathfrak{p}), f(x) /) \tau$, and $e^+_{x}=\exp(\mathfrak{se}^+(\mathfrak{p}),f(x)) / \tau$, where $\tau$ is the temperature parameter. Referring to~(\ref{eq6}), the NCL loss function is:
\begin{equation}
\begin{aligned}
&L_{NCL}=\frac{1}{m \times j}\sum_m\sum_j\log\frac{e_{p_m}^{p_j}}{e_{p_m}^{p_j}+\sum_ie_{p_m}^{n_i}}+\\
&\frac{1}{j}\sum_j\log\frac{e^+_{p_j}}{e^+_{p_j}+\sum_i e^+_{n_i}}+
\frac{1}{i}\sum_i\log\frac{e^-_{n_i}}{e^-_{n_i}+\sum_j e^-_{p_j}}
\end{aligned}
\label{eq7}
\end{equation}

\par      
Last but not least, we explored the exact place to impose the pivot network. Some previous works~\cite{makes2020chai} truncate the deep CNNs at different layers and reveal that the earlier truncated networks provide better features for forgery detection. Besides, the early truncated network has a shallow layout, small reception fields, and a large feature map, which ideally meets the requirement of a small patch size in NCL. Then, we divide the ResNet101 into convolution blocks as in their paper~\cite{ResNet} and explore the feature maps yielded by each ResNet101 block. As expected, the experimental results verify the feature map after the first block to be the most suitable one.  In the Experiments section, we provide more detailed information about the selection of image patch size for NCL.

\subsection{Pivot-consistent Loss}
The pivot network applies convolution on concatenated contour patches; it corrupts the spatial correlations within and among contour patches.~\cite{Hu2020SPAN} has shown spatial information is vital in IML. Therefore, we develop a Pivot-Consistent (PC) loss on the decoder side to ensure that contour patches' spatial correlation remains after the pivot network. PC loss assigns extra weights $\mu$ to contour pixels in the basic pixel-wise BCE loss to enforce the spatial connection among contour pixels. However, the number of contour pixels is far less than the manipulated or authentic pixels. To avoid overfitting, as depicted by ocher arrows on the decoder side in Figure~\ref{fig:onecol} (a), we employ auxiliary classifiers~\cite{cun2020defocus} to accumulate PC loss through each upsampling process gradually. After each upsampling, we shrink the ground truth to the same size as the feature map; pixel-wise IML supervision can then be imposed through shrunken forgery masks. We slightly abuse notations of lower-case letters here. Denote $t$ as pixels in an image, $\hat{t}$ as contour pixels, and $\mu$ as the extra weight. $\gamma(\cdot)$ is the ground truth label for a pixel, $\theta(\cdot)$ is the predicted label of our network for a pixel. $\gamma(\cdot)$ and $\theta(\cdot)$ give binary value as output. Then, our PC loss is:

\begin{equation}
\centering
 \begin{aligned}
 &L_{PC}=\frac{\mu}{\hat{t}}\sum_{\hat{t}} (\gamma({\hat{t}})\log(\theta({\hat{t}}))+(1-\gamma({\hat{t}}))\log(1-\theta({\hat{t}})))\\
 &+\frac{(1-\mu)}{t}\sum_{t} (\gamma(t)\log(\theta(t))+(1-\gamma(t))\log(1-\theta(t)))
  \end{aligned}
 \label{eq8}
\end{equation}
We find larger $\mu$ benefits the final IML accuracy. The assessment of $\mu$ is detailed in the Experiments section.

\subsection{Total Loss Function}
To sum up, NCL for IML has a hybrid total loss as:
\begin{equation}
\begin{aligned}
 L_{total} =\omega \times L_{NCL}+L_{PC} \end{aligned}
\label{eq9}
\end{equation}
$\omega$ is the weight parameter for the non-mutually exclusive contrastive learning on the shallow encoder layers. More of $\omega$ can be found in the Experiments section.

 \section{Experiments and Discussions}
\textbf{Datasets}.
Unlike existing baseline models, our proposed NCL only utilizes four benchmark datasets for training and evaluation. \textbf{No other datasets are involved in our training process}. We train our NCL model on the training split of a dataset and then test it on the corresponding test splite. To distinguish from pre-training, we term our training process conducted only on the benchmark training split as~\textit{benchmark-training}. Our model applies benchmark training in the experiments; unless otherwise stated. The five public datasets for benchmark training and evaluation are: (1) \textbf{CASIA}~\cite{CASIA2013}; (2) \textbf{NIST16}~\cite{NIST2016}; (3) \textbf{Columbia}~\cite{Columbia2009}; (4) \textbf{Coverage}~\cite{COVERAGE2016}; (5)\textbf{Defacto}~\cite{mahfoudi2019defacto}. Training and testing splits of datasets follow the widely accepted practices in~\cite{Wu2019ManTraNet}. For Defacto, the Defacto-84K is used for training and Defacto-12K is applied for testing. In particular, our method does not engage additional datasets, so we follow the standard splits of Coverage, where 75 samples are for training and the rest for testing.

\par
\textbf{Implementation Details}.
As demonstrated in Figure~\ref{fig:onecol} (a), we follow the standard settings of DeepLabV3+ to build the basic encoder-decoder. We adopt an ASPP block with atrous rates of 1, 12, 24, and 36. The $output stride$ is set to 8. The decoder expands encoded features by a factor of 4 until reaching the same size as the input image. We also follow the training protocols in~\cite{DeepLabV3+} to train our proposed model. In detail, we set the batch size to 4 on each dataset. The crop size is 512 $\times$ 512. We adopt Stochastic Gradient Descent (SGD) optimizer with the learning rate schedule ``poly" policy (initial learning rate 0.007, momentum 0.9, and weight decay 5e-4). Our proposed model is trained end-to-end without staged pre-training of each component. Moreover, our total loss is backpropagated as a whole. The weight of NCL loss ($\omega$ in equation~(\ref{eq9})) is 0.01. The weight in PC loss ($\mu$ in equation~(\ref{eq8})) is 0.9. These parameters are set-still in the evaluation.

\textbf{Evaluation Metrics}.
Following the widely accepted practices, we adopt pixel-level $F_1$ score and Area Under the receiver operating characteristic Curve (AUC) as our evaluation metrics. $F_1$ and AUC measure the binary classification accuracy for every pixel. Both metrics range in $[0,100]$ percentage, and higher scores indicate better performances. According to our observation, $F_1$ is more faithful in reflecting the model performance since the numbers of tampered and authentic pixels are extremely unbalanced. AUC will be affected by the huge amount of true-negatives and the optimized AUC threshold will over-estimating the model performance.

\subsection{Quantitative Analysis on Benchmarks}
We compare our model performance with current SoTA methods, including ELA~\cite{ELA2007}, NOI~\cite{NOI2009}, CFA~\cite{CFA2012}, J-LSTM~\cite{J-LSTM2017ICCV}, RGB-N~\cite{zhou2018RGB-N}, ManTra-Net~\cite{Wu2019ManTraNet}, SPAN~\cite{Hu2020SPAN}, OSN~\cite{wu2022robust}, ObjectFormer~\cite{wang2022objectformer}, MVSS~\cite{chen2021image}, and MVSS++~\cite{dong2022mvss} on the five standard datasets. ELA, NOI, and CFA are traditional methods based on hand-crafted features. The rest are end-to-end models. The results measured by the $F_1$ score and AUC are listed respectively in Table~\ref{tb2}. Except for our model, all the other end-to-end methods use considerable additional images for pre-training and benchmark training splits for fine-tuning.

\begin{table*}[h]
\centering
\caption{$F_1$ score (\%) and AUC (\%) comparisons between our proposed method and baselines on benchmarks. }
\setlength{\tabcolsep}{1mm}{
\begin{tabular}{lcccccccccccc}
\hline
 {\multirow{2}{*}{Method}} &  {\multirow{2}{*}{pre-train}} &{\multirow{2}{*}{fine tune}} &
 \multicolumn{2}{c}{NIST16} & \multicolumn{2}{c}{CASIA} &
 \multicolumn{2}{c}{Coverage} & \multicolumn{2}{c}{Columbia}  & \multicolumn{2}{c}{Defacto}\\
 \cline{4-5}  \cline{6-7}   \cline{8-9}  \cline{10-11} \cline{12-13}
&{} &{} &{$F_1$ $\uparrow$ } & {AUC $\uparrow$} & {$F_1$ $\uparrow$} & {AUC$\uparrow$} & {$F_1$ $\uparrow$} & {AUC$\uparrow$} & {$F_1$ $\uparrow$} & {AUC$\uparrow$}  & {$F_1$ $\uparrow$} & {AUC $\uparrow$}\\
\hline
ELA~\cite{ELA2007}  & $\times$ & $\times$     & 23.6 & 42.9   & 21.4 & 61.3  & 22.2 & 58.3  & 47.0 & 58.1 & -& -\\
NOI~\cite{NOI2009}  & $\times$ & $\times$  & 28.5 & 48.7   & 26.3 & 61.2  & 26.9 & 58.7  & 57.4 & 54.6 & -& -\\
CFA~\cite{CFA2012} & $\times$  & $\times$  & 17.4 & 50.1   & 20.7 & 52.2  & 19.0 & 48.5  & 46.7 & 72.0 & -& -\\
\hline
J-LSTM~\cite{J-LSTM2017ICCV}  &\checkmark  &$\times$ & -  & 76.4     & - & -        & - & 61.4     & -&- & -& -\\
ManTra~\cite{Wu2019ManTraNet} &\checkmark &\checkmark  & -  & 79.5     & - & 81.7     & - & 81.9     & -&82.4 & -& -\\
RGB-N~\cite{zhou2018RGB-N}    &\checkmark &\checkmark  & 72.2 & 93.7   & 40.8 & 79.5  & 43.7 & 81.7  & 69.7&85.8& -& -\\
SPAN (1)~\cite{Hu2020SPAN}    &\checkmark &$\times$     & 29.0 & 83.6
& 33.6 & 81.4  & 53.5 & 91.2  & 81.5 &93.6 & -\\
SPAN (2)~\cite{Hu2020SPAN}    &\checkmark &\checkmark  & 58.2 & 96.1
& 38.2 & 83.8  & 55.8 & 93.7  & -&-& -& -\\
ObjectFormer~\cite{wang2022objectformer}    &\checkmark  &\checkmark &82.4 &\textbf{99.6}
 &57.9 &  \textbf{88.2}   &75.8 &95.7  & - & - & -& -\\
OSN~\cite{wu2022robust}  &\checkmark&\checkmark  &28.6 &76.4
 &40.5 &  83.3  &72.7  &88.3  & - & - & -& -\\
MVSS-Net~\cite{chen2021image}    &\checkmark &\checkmark   &- &73.7
 & - & 75.3   & - & 82.4  & - & 72.6 &-  & 53.8\\
MVSS-Net++~\cite{dong2022mvss}    &\checkmark &\checkmark  &-&71.5
& - & 77.1   &- & 52.5    & - & 56.3 &- & 88.6\\

\hline
ours (NCL)   & $\times$  & $\times$ & \textbf{83.1} & 91.2
& \textbf{59.8} & 86.4   & \textbf{80.1} & \textbf{92.8}   & \textbf{85.0} & \textbf{94.3}  & \textbf{60.7} & \textbf{88.9}\\
\hline
\end{tabular}}
\begin{tablenotes}
    \small
    \item  SPAN (1) is under the pre-training setup while SPAN (2) is under the fine-tuning.
    \item  MVSS-Net++ is pre-trained on the Defacto-84K and MVSS-Net is pre-trained on the CASIAv2. 
    \item  `-' denotes that the result is not available in the literature and'$\uparrow$' indicates that the higher value is better.
\end{tablenotes}
\label{tb2}
\end{table*}
In general, our method achieves state-of-the-art performance compared with existing methods. Except for our NCL, all the other methods use a large-scale, synthesized dataset for pre-training and the five benchmarks for fine-tuning. Notably, our model outperforms the others in $F_1$ score. Compared with AUC, $F_1$ score is more faithful in measuring the real performances of an IML. Prior studies uniformly adopt the optimal threshold for the AUC metric, which adjusts the AUC threshold per model and per test. This threshold adjustment is impractical in daily scenarios and commonly overestimates the model behaviors. Therefore, recent researches all turn into more persuasive $F_1$ score or apply fixed threshold when measuring AUC~\cite{dong2022mvss}. Most existing studies do not public their performances measured by fixed AUC. Also, we cannot re-train these models with their pre-training datasets. Here in Table~\ref{tb2}, we adopt the optimal AUC but explicitly show our $F_1$ score to fully demonstrate the surpassing performance of our NCL. Besides, we can find that our model owns a much smaller gap between the $F_1$ score and the AUC value. This indicates higher robustness to some degree.

\subsection{Generalizability and Robustness}
As we conduct benchmark training and benchmark testing, although we achieved state-of-the-art performance and early stop the training epoch at 70, the generalizability of our model is yet to be verified. In other words, we need to answer:``Does NCL overfit these training data?". To address this foremost model generalizability concern, we conduct experiments by training our model on one dataset and then testing it on another non-homogeneous dataset. The result is shown in Table~\ref{cross_tb1}. We first train our NCL model on the relatively large benchmarks, CASIAv2 and Defacto, then test the trained model on the other benchmarks. Since MVSS-Net adopts the same datasets as pre-training datasets, we employ MVSS-Net for comparison. In the first four rows of Table~\ref{cross_tb1}, under the same settings, our NCL exceeds the pre-training-based MVSS-Net in almost every dataset, but NCL does not require the fine-tune on these datasets or extra hand-crafted feature for auxiliary. Therefore, it is clear that NCL does not overfit the training data. Then, to further investigate the generalizability of NCL, we use the smallest two benchmarks, Coverage and Columbia, for training and testing NCL on larger benchmarks. Table~\ref{cross_tb1} indicates NCL manages to cope with this harsh situation. Besides, we also put all the benchmark training datasets together to form a single training dataset and train NCL on this set to probe its edge performance. As shown in the last row of Table~\ref{cross_tb1}, training on this large dataset, NCL gains surpassing performances on almost every single testing dataset regarding existing models. However, compared to NCL with benchmark training, the AUC score is slightly lowered on the Coverage and Columbia datasets but sharply increased on the other three datasets. Considering the small size of Coverage and Columbia, NCL exchanges sensitivity for specificity, thereby achieving more balanced performances regarding all the testing cases. 

\begin{table}[h]
\centering
\caption{AUC (\%) results for generalizability validation.}
\setlength{\tabcolsep}{1.5mm}{
\scalebox{0.68}{
\begin{tabular}{l|c|ccccc}
\hline
{Method}&{\diagbox{Train}{Test}}
&{NIST16} & {CASIA} &{Coverage} & {Columbia} & {Defacto}\\
\hline
MVSS-Net~\cite{chen2021image} &CASIAv2    &73.7 & 75.3    & 82.4   & 72.6  & 53.8\\
ours (NCL)  &CASIAv2    & 75.6          &86.4   & 81.6         & 66.9         &53.0\\
\hline \hline 
MVSS-Net++~\cite{dong2022mvss} &Defacto   &71.5 & 77.1   & 52.5   & 56.3 & 88.6\\
ours (NCL)  &Defacto & 77.3          & 75.6         & 58.7     & 52.3   &88.9 \\
\hline\hline
ours (NCL)  &Coverage & 58.1  & 54.6 &92.8 & 52.2 & 51.9\\
ours (NCL)  &Columbia & 58.9  & 57.3 &51.5 & 95.3 & 52.6\\
\hline\hline
ours  &All* &95.0  & 88.4   & 91.1 & 92.3  & 90.1\\ 
\hline
\end{tabular}}}
\begin{tablenotes}
    \small
    \item  *: All means putting all the benchmark training datasets together (around 25k images) for training.
\end{tablenotes}
\label{cross_tb1}
\end{table}
\par 
Then, we also conduct robustness tests. Typical robustness experiments are conducted through attacks. Built-in functions are used to attack the images, and IML methods are then applied to identify the tampered areas on the attacked images. The results measured by pixel-wise AUC are shown in Table~\ref{tb3}. Our model achieves satisfying robustness against common attacks. Therefore, in short, our NCL-based IML method retains satisfying generalizability and is robust and resistant to attacks.
\begin{table}
\begin{center}
\caption{Robustness analysis of models on NIST16 datasets.}
\label{tb3}
\begin{tabular}{l|c|c|c}
\hline
{Operations} & {ManTra-Net} & {SPAN} & {Ours}  \\
\hline\hline
None  & 79.5 & 83.6 & \textbf{91.2}  \\
\hline\hline
Resize(0.78x)  & 77.4 & 83.2 & \textbf{85.6}  \\
Resize(0.25x)   & 75.5 & 80.3 & \textbf{83.1}\\
\hline
GaussianBlur(size=3)  & 77.4 & 83.1 & \textbf{84.0}  \\
GaussianBlur(size=5)   & 74.6 & 79.2 & \textbf{80.6}\\
\hline
GaussianNoise($\sigma$=3)   & 67.4 & 75.1 & \textbf{79.5}\\
GaussianNoise($\sigma$=5)   & 58.6 & 67.3 & \textbf{71.4}\\
\hline
JPEGCompress(50)   & 77.9 & 83.6 & \textbf{84.3}\\
JPEGCompress(100)   & 74.4 & 80.7 & \textbf{81.9}\\
\hline
\end{tabular}
\end{center}
\begin{tablenotes}
\vspace{-1em}
    \small
    \item SPAN without fine-tuning is adopted here. $100$ and $50$ are the JEPG compress quality factors.
\end{tablenotes}
\end{table}

\subsection{Qualitative Analysis}
\textbf{Contribution of Each Component}.
Before going further, we first clarify the exact improvements brought by each component of our NCL-based method. Our method is built upon the basic encoder-decoder of DeepLabV3+, termed the \textbf{Base model}. Then, we propose the Pivot structure to conduct non-mutually exclusive contrastive learning; we term it as the \textbf{Base+Pivot model}. Further, we engage with the PC loss to form the entire NCL framework; we term the entire NCL as the \textbf{Base+Pivot+PC model} in this section.

\par
\begin{figure}[htbp]
\begin{center}
\includegraphics[width=0.5\textwidth]{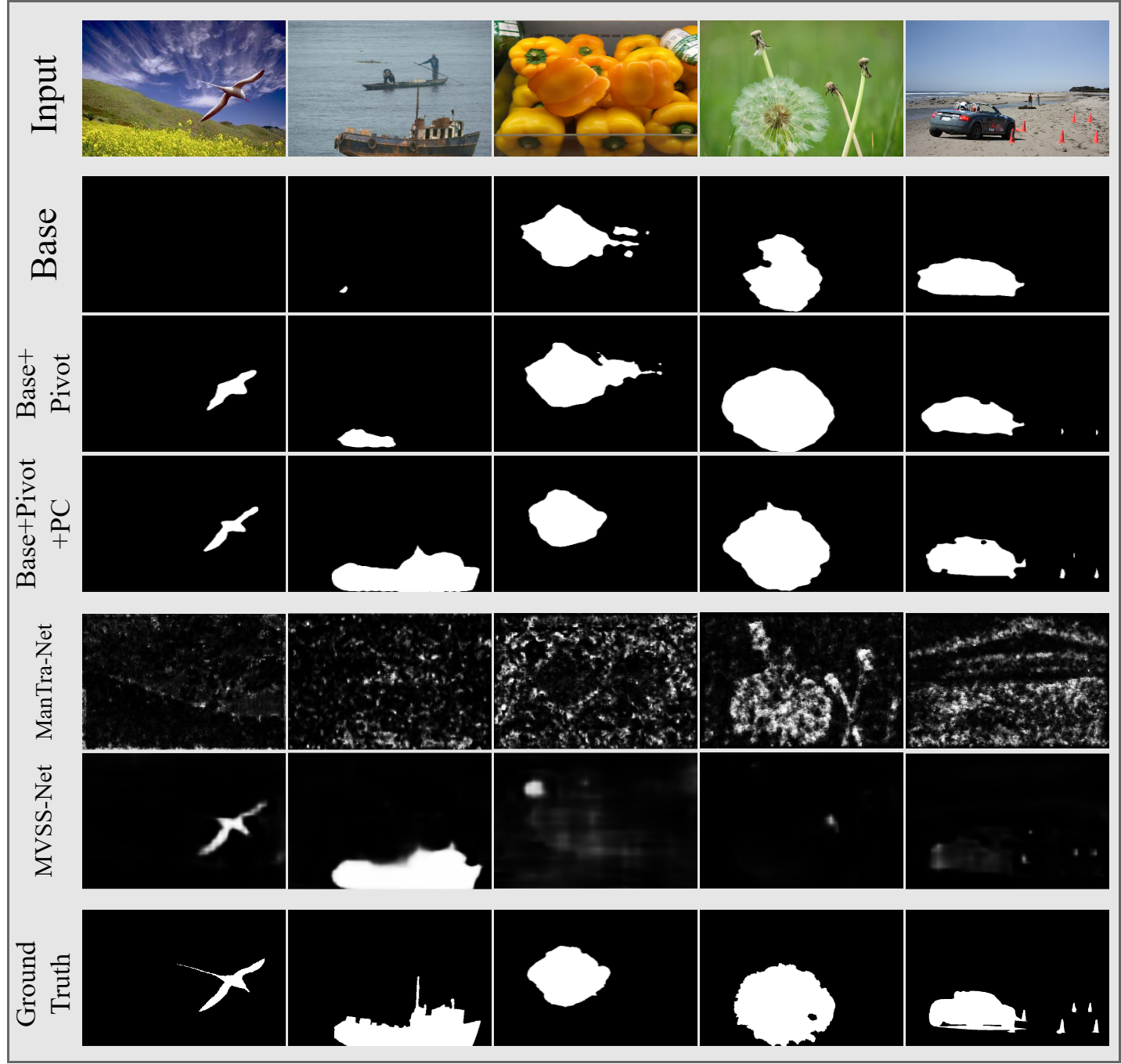}
\end{center}
   \caption{Prediction results of Variants of our methods. From top to bottom: forged images, baseline model, attaching Pivot structure and conducting non-mutually exclusive contrastive learning on the base model, adding PC loss to the former model, Mantra-Net~\cite{Wu2019ManTraNet}, MVSS-Net~\cite{dong2022mvss} and ground-truth masks.}
\label{fig:SQA}
\end{figure}

\begin{figure*}[htbp]
\begin{center}
\includegraphics[width=2.0\columnwidth]{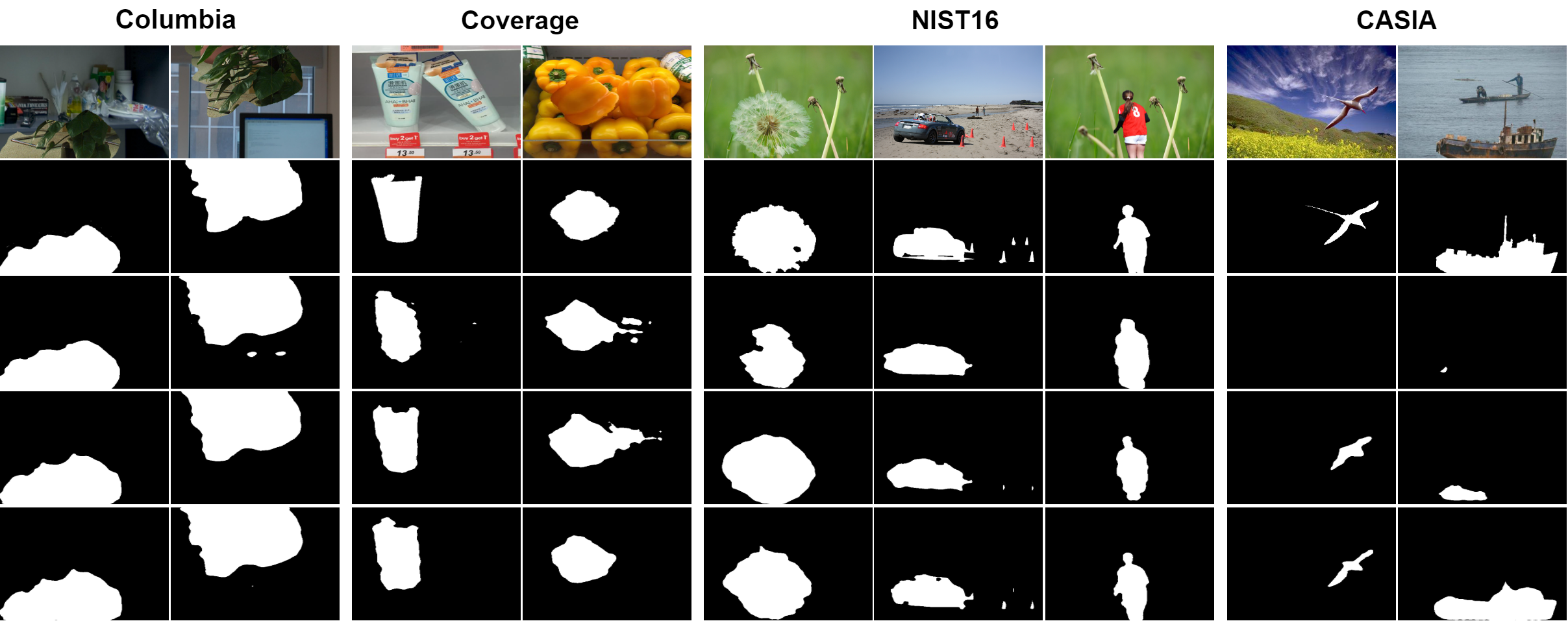}
\end{center}
   \caption{Predictions generated by variations of our model on Columbia, Coverage, NIST16, CASIA datasets. From top to bottom are the: forged images, ground-truth masks, results of Base model, results of Base+Pivot, and results of Base+Pivot+PC.}
\label{fig:SQA2}
\end{figure*}

 
\textbf{Qualitative Analysis}.
We first conduct longitudinal qualitative comparisons of different components among the NCL in Figure~\ref{fig:SQA} and Figure~\ref{fig:SQA2}. The second to fourth rows in Figure~\ref{fig:SQA} are the results of the Base model, Base+Pivot model, Base+Pivot+PC model regarding the input image. The leftmost two columns of Figure~\ref{fig:SQA} vividly demonstrate the efficacy of each component in our method. The Base model totally fails these cases, but Base+Pivot+PC gradually catches the clue of manipulations. The rightmost column and the bell pepper picture present active examples of refining the delicate contour of roughly localized artifacts through PC loss. Shown in Figure~\ref{fig:SQA2}, similar situations are also true in other benchmarks. 
\par 
Then, We conduct horizontal qualitative comparisons of different IML models in the lower-half of Figure~\ref{fig:SQA}. The MVSS-Net and Mantra-Net rows show the corresponding output of the widely compared MVSS-Net and Mantra-Net. With much less and inferior training data, our model outperforms these pre-training-depend and massive-data-required models. 
\par
\par

\begin{table*}[htbp]
\centering
\caption{$F_1$ score (\%) and AUC (\%) comparisons between our proposed method and baselines on benchmarks. }
\label{tb4}
\setlength{\tabcolsep}{1mm}{
\resizebox{2.0\columnwidth}{!}{
\begin{tabular}{lcccccccccccc}
\hline
 {\multirow{2}{*}{Method}} &  {\multirow{2}{*}{pre-train}} &{\multirow{2}{*}{fine tune}} &
 \multicolumn{2}{c}{NIST16} & \multicolumn{2}{c}{CASIA} &
 \multicolumn{2}{c}{Coverage} & \multicolumn{2}{c}{Columbia}  & \multicolumn{2}{c}{Defacto}\\
 \cline{4-5}  \cline{6-7}   \cline{8-9}  \cline{10-11} \cline{12-13}
&{} &{} &{$F_1$ $\uparrow$ } & {AUC $\uparrow$} & {$F_1$ $\uparrow$} & {AUC$\uparrow$} & {$F_1$ $\uparrow$} & {AUC$\uparrow$} & {$F_1$ $\uparrow$} & {AUC$\uparrow$}  & {$F_1$ $\uparrow$} & {AUC $\uparrow$}\\
\hline

ObjectFormer~\cite{wang2022objectformer}   &\checkmark  &\checkmark &82.4 &\textbf{99.6}
 &57.9 &  88.2   &75.8 &95.7  & - & - & -& -\\
ObjectFormer(with NCL appended)   &\checkmark  &\checkmark &\textbf{84.4} &\textbf{99.6}
 &59.1 &  \textbf{88.8}   &77.4 &\textbf{96.0}  & - & - & -& -\\
\hline
ours (NCL)   & $\times$  & $\times$ & 83.1 & 91.2
& \textbf{59.8} & 86.4   & \textbf{80.1} & 92.8   & \textbf{85.0} & \textbf{94.3}  & \textbf{60.7} & \textbf{88.9}\\
\hline
\end{tabular}}
}
\begin{tablenotes}
    \small
    \item  ObjectFormer(with NCL appended) is appending NCL to the output of the ViT in ObjectFormer.
\end{tablenotes}
\end{table*}

\textbf{Quantitative Analysis}.
The prediction results measured in pixel-wise AUC for different variants of our model are shown in Table~\ref{ab2}. Our NCL penetrates significant promotions to the basic encoder-decoder network, especially on model generalization. For base model training on NIST16 or Defacto dataset but testing on other datasets, it fails to generalize on other datasets. Adding the Pivot network to the base model drastically boosts the model generalizability. Base+Pivot gains decent AUC results when testing on different, non-homogeneous datasets. PC loss is also verified to be effective in improving performance. We offer more details in the supplementary materials. After quantifying the contribution of each component, we further probe the effect of the parameters in our modes. We have two parameters, the image patch sizes and weight parameters on the total loss.

\begin{table}[h]
\centering
\caption{AUC (\%) on benchmarks for variants of our model.}
\setlength{\tabcolsep}{1.5mm}{
\scalebox{0.68}{
\begin{tabular}{l|c|ccccc}
\hline
{Method}&{\diagbox{Train}{Test}}
&{NIST16} & {CASIA} &{Coverage} & {Columbia} & {Defacto}\\
\hline
Base  &NIST16    & 75.6          &26.4   & 21.6         & 16.9         &4.3 \\
Base+Pivot &NIST16   & 85.6      &60.8   & 71.7         & 66.1         &41.4    \\
Base+Pivot+PC &NIST16 &91.2      & 65.6  &76.4          & 71.5         &55.0 \\
\hline\hline
Base  &Defacto    & 30.3          &25.8   & 17.4         & 14.0         &54.6 \\
Base+Pivot &Defacto   & 70.4      &72.1   & 51.1         & 50.9         &78.2    \\
Base+Pivot+PC &Defacto & 77.3     &75.6   & 58.7         & 52.3         &88.9 \\
\hline
\end{tabular}}}
\label{ab2}
\end{table}


\par 
\textbf{Image Patch Size}. 
Different encoder layers generate features in different patch sizes, which are vital for the performance of NCL. To find the best patch size, we also try to add the Pivot network after different blocks of ResNet101. In detail, we divide the original ResNet101 into five stages as the original paper~\cite{ResNet} and append the Pivot network at the end of each stage. As shown in Table~\ref{tb6}, the earlier layers outperform the deeper layer a lot, which also confirms the observation from \cite{makes2020chai}. This finding is consistent across benchmarks.

\par 
\textbf{Weight Parameters}.
We explore different allocations of weights to maximize the $F_1$ score and AUC. Under this circumstance, we find the allocation scheme and adopt these parameters as stated in the implementation details. Lower $\omega$ and higher $\mu$ facilitate the IML accuracy in both $F_1$ and AUC. The weight effect is similar across datasets. Thereby, our weight choice is consistent. 

\par 
\textbf{Backbone Architectures}.
With the fast advances in Transformer-based image backbones, we will indeed embrace an IML backbone built on the self-attention mechanism. Like CNNs, ViT~\cite{dosovitskiy2020image} also processes image patches by patches. Therefore, the initial assumption of our NCL holds. Regardless of the patching methods in ViT, the image patches will still be divided into three categories: tampered, authentic, and contour patches. Then, our NCL can be quickly adapted to the ViT-based backbone without any effort and boost the base model's performance. As shown in Table~\ref{tb4}, we did some preliminary tests using the backbone of ObjectFormer~\cite{wang2022objectformer}, and the results met our expectations. 
\par

\begin{table}
\begin{center}
\caption{Performance of our model with Pivot network imposed after different encoder blocks. CASIA dataset is applied.}
\label{tb6}
\begin{tabular}{l|c|c}
\hline
{Pivot network after} & {$F_1$ score} & {AUC}  \\
\hline\hline
 ResNet block 5 & 43.8 & 70.2 \\
 ResNet block 2 & 56.3 & 79.0 \\
 ResNet block 1 & 59.8 & 86.4  \\
\hline
\end{tabular}
\end{center}
\end{table}






\section{Conclusion}
\label{sec:conclusion}
This paper proposes a novel Non-mutually exclusive Contrastive Learning (NCL) paradigm to localize image manipulation without additional pre-training datasets. Our NCL-based IML model reaches state-of-the-art performance, top model generalization, and robustness in all five benchmarks, which indicates our NCL is more applicable to real-life scenarios. To a greater extent, NCL provides a brand-new self-supervised paradigm to tackle tasks with trisected problem spaces like semantic segmentation.

\section{Acknowledgements}
This work is jointly supported by the ``Key Research Program" (Grant No.2022YFC3801304), the Ministry of Science and Technology, PRC, and the ``Fundamental Research Funds for the Central Universities" (Grant No.2022SCU12072, No.YJ2021159). The numerical calculation in this paper has been done at Hefei advanced computing center. The author would like to deliver special thanks to Miss. Chunfang Yu, for her attentive work on the dataset preparation.

{\small
\bibliographystyle{ieee_fullname}
\bibliography{egbib}

\begin{thebibliography}{10}\itemsep=-1pt

\bibitem{NIST2016}
Nist: Nist nimble 2016 datasets.
\newblock \url{https://www.nist.gov/itl/iad/mig/}, 2016.

\bibitem{J-LSTM2017ICCV}
J.~H. {Bappy}, A.~K. {Roy-Chowdhury}, J. {Bunk}, L. {Nataraj}, and B.~S. {Manjunath}.
\newblock Exploiting spatial structure for localizing manipulated image regions.
\newblock In {\em 2017 IEEE International Conference on Computer Vision (ICCV)}, pages 4980--4989, 2017.

\bibitem{makes2020chai}
Lucy Chai, David Bau, Ser-Nam Lim, and Phillip Isola.
\newblock What makes fake images detectable? understanding properties that generalize, 2020.

\bibitem{DeepLabV3+}
Liang~Chieh Chen, Yukun Zhu, George Papandreou, Florian Schroff, and Hartwig Adam.
\newblock Encoder-decoder with atrous separable convolution for semantic image segmentation.
\newblock 2018.

\bibitem{chen2021image}
Xinru Chen, Chengbo Dong, Jiaqi Ji, Juan Cao, and Xirong Li.
\newblock Image manipulation detection by multi-view multi-scale supervision.
\newblock In {\em Proceedings of the IEEE/CVF International Conference on Computer Vision}, pages 14185--14193, 2021.

\bibitem{chen2020improved}
Xinlei Chen, Haoqi Fan, Ross Girshick, and Kaiming He.
\newblock Improved baselines with momentum contrastive learning.
\newblock {\em arXiv preprint arXiv:2003.04297}, 2020.

\bibitem{cun2020defocus}
Xiaodong Cun and Chi-Man Pun.
\newblock Defocus blur detection via depth distillation, 2020.

\bibitem{Deng2020CVPR}
Yu Deng, Jiaolong Yang, Dong Chen, Fang Wen, and Xin Tong.
\newblock Disentangled and controllable face image generation via 3d imitative-contrastive learning.
\newblock In {\em IEEE/CVF Conference on Computer Vision and Pattern Recognition (CVPR)}, June 2020.

\bibitem{dong2022mvss}
Chengbo Dong, Xinru Chen, Ruohan Hu, Juan Cao, and Xirong Li.
\newblock Mvss-net: Multi-view multi-scale supervised networks for image manipulation detection.
\newblock {\em IEEE Transactions on Pattern Analysis and Machine Intelligence}, 2022.

\bibitem{CASIA2013}
Jing Dong, Wei Wang, and Tieniu Tan.
\newblock Casia image tampering detection evaluation database.
\newblock In {\em 2013 IEEE China Summit and International Conference on Signal and Information Processing}, pages 422--426, 2013.

\bibitem{dosovitskiy2020image}
Alexey Dosovitskiy, Lucas Beyer, Alexander Kolesnikov, Dirk Weissenborn, Xiaohua Zhai, Thomas Unterthiner, Mostafa Dehghani, Matthias Minderer, Georg Heigold, Sylvain Gelly, et~al.
\newblock An image is worth 16x16 words: Transformers for image recognition at scale.
\newblock {\em arXiv preprint arXiv:2010.11929}, 2020.

\bibitem{CFA2012}
Pasquale Ferrara, Tiziano Bianchi, Alessia De~Rosa, and Alessandro Piva.
\newblock Image forgery localization via fine-grained analysis of cfa artifacts.
\newblock {\em IEEE Transactions on Information Forensics \& Security}, 7(5):1566--1577, 2012.

\bibitem{gao2021tbnet}
Zan Gao, Chao Sun, Zhiyong Cheng, Weili Guan, Anan Liu, and Meng Wang.
\newblock Tbnet: Two-stream boundary-aware network for generic image manipulation localization.
\newblock {\em arXiv preprint arXiv:2108.04508}, 2021.

\bibitem{he2020momentum}
Kaiming He, Haoqi Fan, Yuxin Wu, Saining Xie, and Ross Girshick.
\newblock Momentum contrast for unsupervised visual representation learning.
\newblock In {\em Proceedings of the IEEE/CVF conference on computer vision and pattern recognition}, pages 9729--9738, 2020.

\bibitem{ResNet}
K. {He}, X. {Zhang}, S. {Ren}, and J. {Sun}.
\newblock Deep residual learning for image recognition.
\newblock In {\em 2016 IEEE Conference on Computer Vision and Pattern Recognition (CVPR)}, pages 770--778, 2016.

\bibitem{Hu2020SPAN}
Xuefeng Hu, Zhihan Zhang, Zhenye Jiang, Syomantak Chaudhuri, Zhenheng Yang, and Ram Nevatia.
\newblock Span: Spatial pyramid attention network forimage manipulation localization.
\newblock 2020.

\bibitem{Self-Consistency2018eccv}
Minyoung Huh, Andrew Liu, Andrew Owens, and Alexei~A. Efros.
\newblock Fighting fake news: Image splice detection via learned self-consistency.
\newblock 2018.

\bibitem{DOA-GAN}
A. {Islam}, C. {Long}, A. {Basharat}, and A. {Hoogs}.
\newblock Doa-gan: Dual-order attentive generative adversarial network for image copy-move forgery detection and localization.
\newblock In {\em 2020 IEEE/CVF Conference on Computer Vision and Pattern Recognition (CVPR)}, pages 4675--4684, 2020.

\bibitem{splice2019MAG}
Vladimir~V. Kniaz, Vladimir~A. Knyaz, and Fabio Remondino.
\newblock The point where reality meets fantasy: Mixed ` adversarial generators for image splice detection.
\newblock In {\em Advances in Neural Information Processing Systems 32: Annual Conference on Neural Information Processing Systems 2019, NeurIPS 2019, December 8-14, 2019, Vancouver, BC, Canada}, pages 215--226, 2019.

\bibitem{ELA2007}
N. Krawetz.
\newblock A picture’s worth...
\newblock {\em Hacker Factor Solutions}, 6(2), 2007.

\bibitem{BLK2009}
Weihai Li, Yuan Yuan, and Nenghai Yu.
\newblock Passive detection of doctored jpeg image via block artifact grid extraction.
\newblock {\em Signal Processing}, 89(9):1821--1829, 2009.

\bibitem{ma2023iml}
Xiaochen Ma, Bo Du, Zhuohang Jiang, Ahmed Y.~Al Hammadi, and Jizhe Zhou.
\newblock Iml-vit: Benchmarking image manipulation localization by vision transformer, 2023.

\bibitem{NOI2009}
Babak Mahdian and Stanislav Saic.
\newblock Using noise inconsistencies for blind image forensics.
\newblock {\em Image and Vision Computing}, 2009.

\bibitem{mahfoudi2019defacto}
Ga{\"e}l Mahfoudi, Badr Tajini, Florent Retraint, Frederic Morain-Nicolier, Jean~Luc Dugelay, and PIC Marc.
\newblock Defacto: Image and face manipulation dataset.
\newblock In {\em 2019 27th European Signal Processing Conference (EUSIPCO)}, pages 1--5. IEEE, 2019.

\bibitem{mayer2019forensic}
Owen Mayer and Matthew~C Stamm.
\newblock Forensic similarity for digital images.
\newblock {\em IEEE Transactions on Information Forensics and Security}, 15:1331--1346, 2019.

\bibitem{Columbia2009}
T Ng, T, J Hsu, and S Chang.
\newblock Columbia image splicing detection evaluation dataset.
\newblock In {\em DVMM lab. Columbia Univ CalPhotos Digit Libr}, 2009.

\bibitem{robinson2020contrastive}
Joshua Robinson, Ching-Yao Chuang, Suvrit Sra, and Stefanie Jegelka.
\newblock Contrastive learning with hard negative samples.
\newblock {\em International Conference on Learning Representations}, 2021.

\bibitem{splice2017MFCN}
Ronald Salloum, Yuzhuo Ren, and C.~C.~Jay Kuo.
\newblock Image splicing localization using a multi-task fully convolutional network (mfcn).
\newblock {\em Journal of Visual Communication \& Image Representation}, 51(feb.):201--209, 2017.

\bibitem{wang2022objectformer}
Junke Wang, Zuxuan Wu, Jingjing Chen, Xintong Han, Abhinav Shrivastava, Ser-Nam Lim, and Yu-Gang Jiang.
\newblock Objectformer for image manipulation detection and localization.
\newblock In {\em Proceedings of the IEEE/CVF Conference on Computer Vision and Pattern Recognition}, pages 2364--2373, 2022.

\bibitem{COVERAGE2016}
Bihan Wen, Ye Zhu, Ramanathan Subramanian, Tian~Tsong Ng, and Stefan Winkler.
\newblock Coverage — a novel database for copy-move forgery detection.
\newblock In {\em IEEE International Conference on Image Processing}, 2016.

\bibitem{wu2022robust}
Haiwei Wu, Jiantao Zhou, Jinyu Tian, and Jun Liu.
\newblock Robust image forgery detection over online social network shared images.
\newblock In {\em Proceedings of the IEEE/CVF Conference on Computer Vision and Pattern Recognition}, pages 13440--13449, 2022.

\bibitem{spliceDT2017}
Yue Wu, Wael Abd-Almageed, and Prem Natarajan.
\newblock Deep matching and validation network: An end-to-end solution to constrained image splicing localization and detection.
\newblock pages 1480--1502, 10 2017.

\bibitem{BusterNet}
Y. Wu, W. Abdalmageed, and P. Natarajan.
\newblock Busternet: Detecting copy-move image forgery with source/target localization.
\newblock In {\em European Conference on Computer Vision (ECCV)}, 2018.

\bibitem{Wu2019ManTraNet}
Y. {Wu}, W. {AbdAlmageed}, and P. {Natarajan}.
\newblock Mantra-net: Manipulation tracing network for detection and localization of image forgeries with anomalous features.
\newblock In {\em 2019 IEEE/CVF Conference on Computer Vision and Pattern Recognition (CVPR)}, pages 9535--9544, 2019.

\bibitem{xie2021propagate}
Zhenda Xie, Yutong Lin, Zheng Zhang, Yue Cao, Stephen Lin, and Han Hu.
\newblock Propagate yourself: Exploring pixel-level consistency for unsupervised visual representation learning.
\newblock In {\em Proceedings of the IEEE/CVF Conference on Computer Vision and Pattern Recognition}, pages 16684--16693, 2021.

\bibitem{zhou2020personal}
Jizhe Zhou and Chi-Man Pun.
\newblock Personal privacy protection via irrelevant faces tracking and pixelation in video live streaming.
\newblock {\em IEEE Transactions on Information Forensics and Security}, 16:1088--1103, 2020.

\bibitem{zhou2020privacy}
Jizhe Zhou, Chi-Man Pun, and Yu Tong.
\newblock Privacy-sensitive objects pixelation for live video streaming.
\newblock In {\em Proceedings of the 28th ACM International Conference on Multimedia}, pages 3025--3033, 2020.

\bibitem{zhou2018RGB-N}
P. {Zhou}, X. {Han}, V.~I. {Morariu}, and L.~S. {Davis}.
\newblock Learning rich features for image manipulation detection.
\newblock In {\em 2018 IEEE/CVF Conference on Computer Vision and Pattern Recognition}, pages 1053--1061, 2018.

\bibitem{zhuo2021self}
Long Zhuo, Shunquan Tan, Bin Li, and Jiwu Huang.
\newblock Self-adversarial training incorporating forgery attention for image forgery localization.
\newblock {\em arXiv preprint arXiv:2107.02434}, 2021.

\end{thebibliography}
}

\end{document}